%% file: main.tex
\documentclass{article} % For LaTeX2e
\usepackage{iclr2025_conference,times}
\iclrfinalcopy

\input{math_commands.tex}

\usepackage{hyperref}
\usepackage{url}

\usepackage{times}
\usepackage{latexsym}
\usepackage[T1]{fontenc}
\usepackage[utf8]{inputenc}
\usepackage{microtype}
\usepackage{inconsolata}
\usepackage{graphicx}
\usepackage{algorithm}
\usepackage{algpseudocode}
\usepackage{amssymb}
\usepackage{mathtools}
\usepackage{amsthm}

\usepackage{arydshln}
\usepackage{listings}
\usepackage{multirow}
\usepackage{multicol}
\usepackage{color, colortbl}
\usepackage{tcolorbox}
\usepackage{xcolor}
\usepackage{amsmath}
\usepackage{enumitem}
\usepackage{etoolbox,lipsum}
\usepackage{subcaption}
\usepackage{stfloats}
\usepackage{float}

\definecolor{LightCyan}{rgb}{0.93,1,1}
\definecolor{LLightCyan}{rgb}{0.97,1,1}
\definecolor{LightRed}{rgb}{1,0.95,0.95}
\definecolor{LLightRed}{rgb}{1,0.97,0.97}
\definecolor{Gray}{rgb}{0.85,0.85,0.85}
\definecolor{LightGray}{rgb}{0.8,0.8,0.8}
\definecolor{LLightRed}{rgb}{1,0.93,0.95}
\definecolor{LLightBlue}{rgb}{0.9,0.95,1}
\definecolor{LLightGreen}{rgb}{0.95,1,0.98}
\definecolor{SVRAG}{rgb}{1,1,0.95}
\definecolor{LLightGray}{rgb}{0.95,0.95,0.95}
\definecolor{LLLightGray}{rgb}{0.98,0.98,0.98}
\definecolor{mygreen}{rgb}{0.4,0.7,0.306}
\definecolor{myblue}{rgb}{0.294,0.447,0.796}
\usepackage{hhline}

\definecolor{Gray}{gray}{0.85}
\definecolor{LightCyan}{rgb}{0.88,1,1}

\usepackage[textsize=tiny]{todonotes}

\DeclareRobustCommand{\papername}{SV-RAG}
\DeclareRobustCommand{\dataname}{VisR}

\title{\papername: LoRA-Contextualizing Adaptation of \\ MLLMs for Long Document Understanding}

\author{Jian Chen$^{1}$\thanks{Equal contribution, work done when JC is at Adobe Research.}~, Ruiyi Zhang$^{2*}$, Yufan Zhou$^{2}$, Tong Yu$^{2}$, \textbf{Franck Dernoncourt}$^{2}$\\
\textbf{Jiuxiang Gu}$^{2}$, \textbf{Ryan Rossi}$^{2}$, \textbf{Changyou Chen}$^{1}$, \textbf{Tong Sun}$^{2}$ \\
University at Buffalo$^1$, Adobe Research$^{2}$\\
\texttt{ruizhang@adobe.com}
}

\IfFileExists{main_backup.aux}{
  \message{We saw a default backup.aux file, let's use it instead of the main aux file.}
  \nofiles % Disable default aux file
  \makeatletter
  \input{main_backup.aux}
  \makeatother
}{}

\begin{document}
\maketitle
\vspace{-1em}
\begin{abstract}
Multimodal large language models (MLLMs) have recently shown great progress in text-rich image understanding, yet they still struggle with complex, multi-page visually-rich documents. Traditional methods using document parsers for retrieval-augmented generation suffer from performance and efficiency limitations, while directly presenting all pages to MLLMs leads to inefficiencies, especially with lengthy ones.  In this work, we present a novel framework named \textbf{S}elf-\textbf{V}isual \textbf{R}etrieval-\textbf{A}ugmented \textbf{G}eneration (SV-RAG
), which can broaden horizons of \textit{any} MLLM to support long-document understanding. We demonstrate that \textbf{MLLMs themselves can be an effective multimodal retriever} to fetch relevant pages and then answer user questions based on these pages. SV-RAG is implemented with two specific MLLM adapters, one for evidence page retrieval and the other for question answering. Empirical results show state-of-the-art performance on public benchmarks, demonstrating the effectiveness of SV-RAG.
\end{abstract}

\input{sections/introduction}
\input{sections/related_work}
\input{sections/method}

\input{sections/experiment}

\section{Conclusions}
In this paper, we propose \papername, a lightweight MLLMs for visually-rich document understanding. \papername~has a unique design to facilitate multi-page document understanding using dual LoRA adapters. The research highlights that small open-source models are great at processing multipage documents and underscored the importance of efficient retrieval mechanisms in filtering irrelevant pages. 
Furthermore, we collect the VisR-bench dataset for document understanding, and empirical results on benchmarks demonstrated the effectiveness of \papername. We hope these findings provide valuable insights for optimizing lightweight MLLMs, aiming to improve accuracy and efficiency in visually-rich document understanding.

\section{Limitations}
\papername~is the first MLLM that can perform visual retrieval-augmented generation for document question answering using a single model. However, it still requires computational resources for training and inference, which may limit its practical applicability in resource-constrained environments. \papername~should be mobile friendly, as it only requires a single base model. This base model can be Phi-3-Silica within MS operating systems or an Apple on-device model within Apple IOS 18. A routing mechanism in Apple Intelligence can better balance computational cost and performance.  However, our experiments are not performed on these real-world devices, which are necessary for pushing forward document intelligence.

\section{Ethics Statement}
The \dataname-Bench dataset was curated with careful consideration of ethical and legal concerns. All documents are sourced from publicly available data with licenses explicitly permitting research use. To ensure data integrity and compliance, we provide links to the original sources instead of distributing the documents. Additionally, all QA pairs have been manually reviewed to exclude harmful content and personally identifiable information. The dataset does not expose sensitive user data, and experimental results are reported as aggregate statistics to prevent information leakage while ensuring reproducibility. These measures uphold ethical and legal standards while supporting responsible AI research.

\bibliographystyle{iclr2025_conference}

\clearpage
\appendix
\input{sections/appendix}

\end{document}

%% file: math_commands.tex
\usepackage{amsmath,amsfonts,bm}

\def\eqref#1{equation~\ref{#1}}
\def\1{\bm{1}}

\DeclareMathAlphabet{\mathsfit}{\encodingdefault}{\sfdefault}{m}{sl}
\SetMathAlphabet{\mathsfit}{bold}{\encodingdefault}{\sfdefault}{bx}{n}

\DeclareMathOperator*{\argmax}{arg\,max}
\DeclareMathOperator*{\argmin}{arg\,min}

%% file: sections/introduction.tex
\section{Introduction}
Documents serve as a critical medium for the preservation and dissemination of information, with millions produced annually. These documents are not limited to simple text; they encompass complex layouts and a variety of modalities such as text, tables, figures, and charts. Visually-rich document understanding (VDU) is thus an essential and challenging area of research. Recently, Multimodal Large Language Models (MLLMs) has emerged, showcasing remarkable abilities to process and understand documents. These models span both proprietary and open-source domains, like GPT-4o \citep{openai2023gpt4}, Gemini-1.5 \citep{team2023gemini}, and Claude-3 among proprietary models, and InternLM-XC2-4KHD \citep{dong2024internlm}, InternVL-Chat \citep{chen2023internvl}, LLaVA-NeXT \citep{liu2024llavanext}, Mini-CPM \citep{hu2024minicpm}, mPLUG-DocOwl \citep{ye2023mplugowl}, LLaVAR~\citep{zhang2023llavar} and TextMonkey \citep{liu2024textmonkey} in open-source space. Their performance has been particularly notable in single-page DU tasks demonstrated on datasets like DocVQA \citep{mathew2021docvqa}, ChartQA \citep{masry2022chartqa} and InfoVQA \citep{mathew2022infographicvqa}.

In real-world applications, they often present documents that are much longer, containing dozens or hundreds of pages\citep{ma2024mmlongbench,tanaka2023slidevqa,islam2023financebench,zhu2021tat}. Addressing the understanding of such lengthy documents presents MLLMs with new challenges \citep{ma2024mmlongbench}. One way is to utilize a classical document parser \citep{rausch2021docparser} to extract information and formulate a prompt for LLM \citep{wang2023layout,lamott2024lapdoc}, which is difficult to recover the layout in prompts and suffers performance degeneration from the document parser. The other way is to exploit the long context windows of large models, allowing them to take multiple pages at once. However, most of the input pages are not relevant to user requests, and efficiency will be compromised when the document contains hundreds of pages~\cite{ma2024mmlongbench,islam2023financebench} or there is a document collection \citep{tito2021document}. 

In this work, we first retrieve evidence pages to obtain relevant information within a vast and varied landscape of content. Unlike using a classical document parser, we propose using MLLMs as the visual retriever, which have shown great generalization ability as they have been trained on a huge text corpus. After obtaining the embedding of each page, we further utilize contextualized late interaction for relevance scoring \citep{khattab2020colbert}. This design shows significantly better efficiency and accuracy than using the classical document parser to extract information. Top-$k$ pages are then selected from hundreds of pages and provided to MLLMs to answer user questions on documents. 

\begin{figure}[!h]
    \centering
    \vspace{-0.5em}
    \includegraphics[width=1\textwidth]{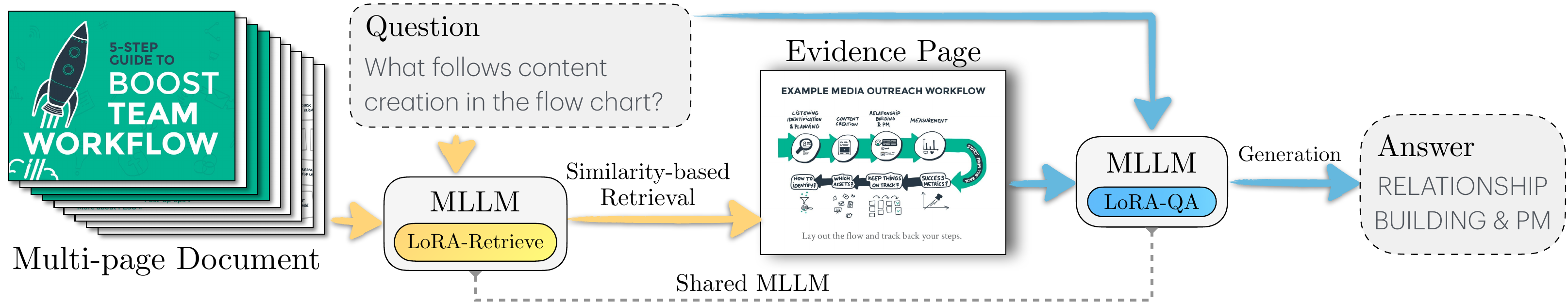}
    \caption{Overview of the \papername~pipeline. The multi-page document and query are encoded by a customized MLLM (yellow). The most relevant page is retrieved through similarity-based matching, and a fine-tuned MLLM (blue) generates the final answer from the evidence.}
    \label{fig:abstract}
    \vspace{-0.5em}
\end{figure}

Based on this design demonstrated in Figure \ref{fig:abstract}, we introduce the \papername~framework for multi-page document understanding, which includes modules for evidence page retrieval and answer generation. Our contributions can be summarized as follows.
\begin{itemize}[itemsep=1pt, topsep=1pt,leftmargin=*]
\item We propose a novel framework named \papername~to broaden the horizons of MLLMs, where we use intermediate MLLMs hidden embedding for \textbf{efficient} question-based evidence page retrieval.  
\item We finetune MLLMs through dual LoRA adapters for evidence page retrieval and question answering, respectively, enabling \papername~to be edge-friendly with great memory efficiency.
\item We collect a visually-rich document QA dataset, \dataname-Bench, comprising nine domains including magazine, flyer, newsletter, product manual, and presentations, etc. This dataset is built upon web-crawl documents, containing 226 documents and 471 question answer pairs.
\item We empirically show that \papername, with only 4B parameters, achieves state-of-the-art performance on \dataname-Bench and four public benchmarks, rivaling Gemini-1.5-pro on MMLongBench-Doc and demonstrating its effectiveness.
\end{itemize}

%% file: sections/related_work.tex
\section{Related Work}

\paragraph{Visually-rich Document Understanding}
Visual Document Understanding (VDU) is the field focused on interpreting text-rich images such as tables \citep{zhong2019image}, charts \citep{masry2022chartqa}, and webpage screenshots \citep{liu2024visualwebbench,tanaka2021visualmrc}. These images are complex, featuring a mix of text and visual elements that convey abundant information \citep{gu2024adopd}. To evaluate multimodal document understanding, tasks range from low-level recognition tasks, such as information extraction, to high-level cognitive tasks, such as visual question answering \citep{mathew2020docvqa}. Models in VDU are typically divided into two categories: OCR-dependent \citep{xu2020layoutlm} and OCR-free, based on their reliance on Optical Character Recognition (OCR). OCR-dependent models are often trained to synchronize textual and visual data. For instance, UDoP \citep{tang2023unifying} is pre-trained to restore obscured textual and layout details using both image and partial text inputs. OCR-free approaches must include text recognition training. Dount \citep{donut} is an example of an OCR-free model that focuses on producing unbroken text sequences, disregarding structural details. In contrast, Pix2Struct \citep{lee2023pix2struct}, another OCR-free model, focuses on interpreting the structure by creating HTML DOM trees from webpage screenshots. However, this technique does not easily transfer to other image types. Our method focuses on the visual question-answering task, specifically targeting questions over long documents consisting of multiple pages of multimodal information.
\vspace{-0.5em}

\paragraph{Multimodal Retrieval-Augmented Generation}
Augmenting language models with information from various knowledge sources has been found to boost their performance in different NLP tasks. The Dense Passage Retriever (DPR) \citep{karpukhin2020dense} trains its retrieval mechanism with documents from within the same batch, using contrastive learning with negatively sampled examples, which enhances its capabilities in open-domain question answering. Document Screenshot Embedding (DSE) \citep{ma2024unifying} uses MLLMs as encoders for both document screenshots and queries, training through contrastive learning to achieve enhanced multimodal retrieval. Both REALM \citep{guu2020retrieval} and Retrieval-Augmented Generation (RAG) \citep{gao2023retrieval} consider the passages they retrieve as hidden variables and train the retrieval and generation components together, improving the efficiency of the retrieval-augmented generation approach. Taking cues from textual RAG, the Plug-and-play \citep{chen2024plug} approach uses GradCAM \citep{selvaraju2020grad} to fetch pertinent image segments corresponding to a given query. The MuRAG \citep{chen2022murag} model introduces a multimodal retrieval-augmented Transformer that utilizes an external multimodal memory for language generation enhancement. Unlike other approaches that retrieve information from various knowledge sources, \papername~focuses on retrieving relevant evidence pages from a given document. This helps MLLMs generate accurate and explainable answers based on the retrieved content. MM-GEM~\cite{ma2024multi} trains an MLLM with a hybrid loss for similarity computation and caption generation on natural images. In contrast, our approach targets visually rich documents, using two LoRA modules to specialize in each task.

\paragraph{Multimodal Large Language Models}
\vspace{-0.5em}
While Large Language Models (LLMs) excel at text-only question answering (QA) \citep{dasigi2021dataset, lee2023qasa}, they cannot process other modalities. To enable multimodal tasks like Visual Question Answering (VQA), MLLMs transform images and videos into visual tokens that LLMs can understand. To train these MLLMs, MiniGPT-4  \citep{zhu2023minigpt4} leverages ChatGPT to produce data compliant with high-quality instructions, while LLaVA \citep{liu2023visual} prompts GPT-4 with image captions and bounding boxes. \citet{chen2023sharegpt4v, chen2024allava} have prompted OpenAI GPT-4V to generate more than 1M pieces of quality data to train MLLMs. LLaMA-Adapter \citep{zhang2023llamaadapter,gao2023llamaadapterv2} and mPLUG-Owl \citep{ye2023mplugowl} align text and image features with large-scale image-text pairs. InstructBLIP \citep{dai2023instructblip} has restructured 13 vision-language tasks to fit an instruction-based approach. mPLUG-Owl \citep{ye2023ureader,ye2023mplugowl} implements multi-task instruction fine-tuning with public document datasets. Recent research \citep{liu2023improvedllava, liu2024llavanext, bai2023qwen, dong2024internlm, xu2024llava, luo2024feast} improves visual encoders by increasing resolution, leading to significant advances in downstream applications but also raising memory costs, especially in multi-page tasks. TextMonkey \cite{liu2024textmonkey} compresses visual tokens using a token resampler. Our method extends MLLMs to handle multi-page documents by retrieving relevant pages, reducing computation and distractions from long token sequences.

%% file: sections/method.tex
\section{\papername~Method}
Multi-page document understanding aims to answer questions related to long and complex documents containing both text and images from users. We denote a document of $n$-pages as a sequence of images, $\mathbf{X}=\{\mathbf{x}_1, \mathbf{x}_2, \dots, \mathbf{x}_n\}$. Text token sequence of a question $q$ is denoted as $\{q_1, q_2, \dots, q_n\}$. Traditional approaches that begin with a parsing step to extract content elements such as images, tables, and forms from documents, then generate answers based on these contents using LLMs~\citep{saad2023pdftriage,wang2023layout}. Here, we first consider using MLLMs to handle this task and avoiding the heuristic document parsing process, where we directly convert each page into a single image. It is not desired, as most pages in a document are irrelevant to user questions and performing an evidence page retrieval can further enhance the efficiency.

We introduce \papername, a method that efficiently leverages the capabilities of pre-trained MLLMs for long document question-answering (QA). \papername~can broaden the horizon of MLLMs to answer questions over long documents or document collections with hundreds of pages. This finding is based on the fact that hidden states of MLLMs can be effective page representations for question-based retrieval, as shown in Section \ref{sec:ablation}. This representation ability can be further enhanced with contrastive training using a LoRA adapter, demonstrating surprising retrieval performance of MLLMs. Furthermore, we can finetune a LoRA-adpter of QA to further enhance the performance of \papername~on specific domains. In summary, we first retrieve evidence pages to rank these images based on their relevance score to a given question $q$, then select the most relevant images, which are then fed into the MLLM to generate the answer. In this section, we introduce the \papername~architecture in Section \ref{sec:architecture}, retrieval training in Section \ref{sec:col} and dual-adapter designs in Section \ref{sec:dual_adapter}.

\begin{figure}[!h]
    \centering
    \includegraphics[width=0.95\textwidth]{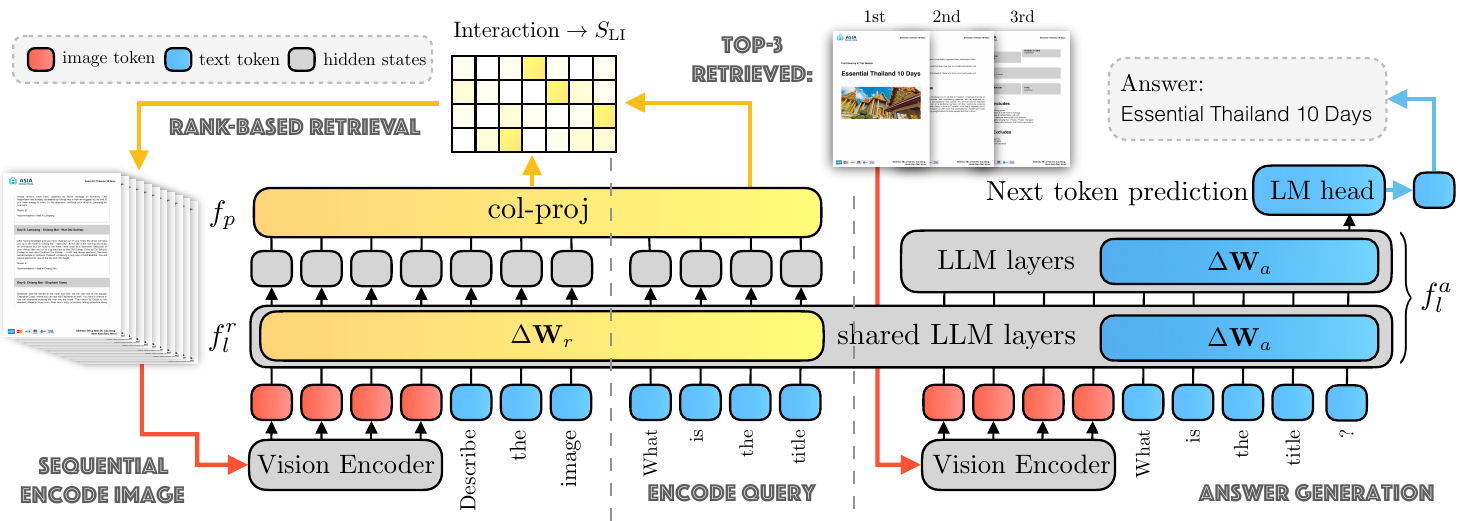}
    % \vspace{-1em}
    \caption{Model overview of \papername. It contains two modules, which are finetuned using LoRA~\citep{hu2021lora}, sharing the \textbf{same} pretrained multimodal LLM backbone. The retrieval module selects evidence pages for the other QA module, which provides responses to user questions.}
    \label{fig:overview}
    % \vspace{-1em}
\end{figure}

\subsection{Architecture}
\label{sec:architecture}
Figure \ref{fig:overview} presents an overview of our model architecture, which comprises two MLLM-based modules for the retrieval of evidence pages and question answers. 

\paragraph{Col-Retrieval Module} 
Building on the approach introduced in ColPali \citep{faysse2024colpali}, we employ a modified MLLM for retrieval, comprising a vision encoder $f_{v}$, a large language model (LLM) $f_{l}^r$, and a Col-projection layer $f_p$. For an input image $\mathbf{X}$, the vision encoder computes a sequence of visual embeddings $f_{v}(\mathbf{X})$, which are then concatenated with token embeddings $\mathbf{y}_v$ derived from a fixed text prompt: “$\backslash$nDescribe the image.” This combined input is fed into the LLM. The projection layer $f_p$ then transforms the LLM’s last hidden states into a low-dimensional feature space, resulting in feature sequences that can be represented as $\mathbf{E}_v = f_p(f_{l}^r(f_{v}(\mathbf{X}), \mathbf{y}_v))$. Similarly, for an input question $q$, the question is first augmented into $y_q$ using a prompt template. Then, its token embedding $\mathbf{y}_q$ is processed without visual input as $\mathbf{E}_q = f_p(f_{l}^r(\mathbf{y}_q))$. Finally, a late-interaction score $s_{\text{LI}}(\mathbf{E}_q, \mathbf{E}_v)$ is computed between the feature sequences, measuring the relevance of a page image to the question text. More details about scoring method is provided in section \ref{sec:col}.

\paragraph{Question-Answering Module} 
The QA module uses a classic LLaVA-like architecture \citep{liu2024visual}, utilizing a vision encoder $f_{v}$ to compute visual embeddings, which are combined with token embeddings and processed by an LLM $f_{l}^a$. The LLM then generates text answers autoregressively through next-word prediction.

\subsection{Contextualized Late Interaction}
\label{sec:col}
We utilize the contextualized late interaction (Col) technique \citep{khattab2020colbert} to compute relevance scores for evidence retrieval. Unlike traditional single-vector encoders, such as CLIP \citep{radford2021learning}, the Col technique introduces an inter-sequence similarity metric called the late-interaction score, which captures more fine-grained question-image relevance. Formally, the late-interaction score between a text feature sequence $\mathbf{E}_q = \{\mathbf{e}_{q_1}, \dots, \mathbf{e}_{q_n}\}$ of length $n$ and a visual feature sequence $\mathbf{E}_v = \{\mathbf{e}_{v_1}, \dots, \mathbf{e}_{v_m}\}$ of length $m$ is defined as:
\begin{equation}
\label{eq:late_interaction}
% \scalebox{0.95}{$
    s_{\text{LI}}(\mathbf{E}_q, \mathbf{E}_v) = \sum_{i=1}^n \max_{j \in \{1,..,m\}} \mathbf{e}_{q_i} \cdot \mathbf{e}_{v_j}^T.
% $}
\end{equation}
We use it as a similarity score in contrastive learning to facilitate ranked retrieval. Specifically, we train our retrieval module to maximize the late-interaction score between a question and its corresponding evidence image, considering these as positive pairs. We then identify the most similar, but unassociated, image within the batch to form the hardest negative pair and minimize the score for this pair. Figure \ref{fig:negative} shows a training pair example. The loss function is defined as:
\begin{equation}
\label{eq:loss}
% \scalebox{0.9}{$
    \mathcal{L} = \log(1+\exp(s_{\text{LI}}(\mathbf{E}_q, \mathbf{E}_v^{-}) - s_{\text{LI}}(\mathbf{E}_q, \mathbf{E}_v^{+}))).
% $}
\end{equation}
The training process of the Col-retrieval module is summarized in Algorithm \ref{alg:training}.
\begin{algorithm}[!ht]
   \caption{Col-retrieval training}
   \label{alg:training}
\begin{algorithmic}[1]
   \Require Pre-trained MLLM $\{f_v$, $f_l^r\}$, training batch of evidence image and question pairs $\{(\mathbf{X}_1, \mathbf{y}_1),\cdots,(\mathbf{X}_b, \mathbf{y}_b)\}$.
   \State Initialize the Col-projection layer $f_p$.
   \While{not converged}
   \State Get \scalebox{0.9}{$\mathbf{E}_v^i = f_p(f_{l}^r(f_{v}(\mathbf{X}_i), \mathbf{y}_i))$}, \scalebox{0.9}{$i \in \{1,...,b\}$}.
   \State Get \scalebox{0.9}{$\mathbf{E}_q^i = f_p(f_{l}^r(\mathbf{y}_i))$}, \scalebox{0.9}{$i \in \{1,...,b\}$}.
   \State Compute \scalebox{0.9}{$\mathbf{S}_{i,j} = s_{\text{LI}}(\mathbf{E}_q^i, \mathbf{E}_v^j)$}.
   \State Get negative image index $\hat{i}$ for each $\mathbf{y}_i$: $\hat{i}=\argmax_{j\in \{1,...,b\}, j\neq i}(\mathbf{S}_{i,j})$
   \State Gradient update using loss function Eq.(\ref{eq:loss}), \Statex \hspace{\algorithmicindent} where $\mathbf{E}_v^{j-}=\mathbf{E}_{\hat{i}}$.
    \EndWhile
\end{algorithmic}
\end{algorithm}

\subsection{Parameter Sharing via Dual-Adapter}
\label{sec:dual_adapter}
To reduce memory usage, we optimize the model by sharing a single MLLM that includes both the vision encoder $f_v$ and the language model $f_l$ across both the retrieval and QA modules. To accommodate the different tasks required by each module, we insert two sets of adapters into the $f_l$ using the LoRA method \citep{hu2021lora}. In the retrieval module, we use a set of adapters $\Delta \mathbf{W}_r$ to create the retrieval-LLM, $f_{l}^r$. For the QA module, a different set of adapters $\Delta \mathbf{W}_a$ is added to the $f_l$, creating the QA-LLM, $f_{l}^r$. In this way, we support both tasks using a single LLM and vision encoder, adding only $\sim$2\% additional parameters.

%% file: sections/experiment.tex
\section{\dataname-Bench}
\paragraph{Visually-rich Document Selection} About 4,000 PDF documents are crawled from the Web and contents of these documents are extracted via a document parser\footnote{Adobe Extract API: \textcolor{magenta}{\href{https://developer.adobe.com/document-services/docs/overview/pdf-extract-api/}{https://developer.adobe.com/document-services/apis/pdf-extract/}}}.
We keep the document with figures and throw away text-only or scan documents. To select documents with specific types of figures, we build a figure scheme that includes 19 figure types after reviewing different documents. We find some types of figures are not informative, such as logo and banner. We use the pretrained CLIP model \texttt{ViT-L/14-336}~\citep{radford2021learning} to perform a figure classification on the extracted figures and keep 6 out of 19 types of figures, including tables, maps, diagrams, infographics, data charts, workflows, and screenshots. After that, we also annotate the document types for all selected documents. 

\paragraph{Question-Answer Collection} Document parser returns all document elements in JSON format and the figures are saved separately as image files. We retrieve the JSON file for the document to obtain the contexts of each figure. Then we combine the figures with their contexts and use GPT-4o (API version 2024-02-15-preview) to generate QA pairs. For the GPT-4o prompts, we provide two demonstrations and ask GPT-4o to generate a QA pair. In addition, we perform automatic verification using GPT-4o to ensure the quality of the generation. Specifically, we only provide the figure to GPT-4o and ask it with the generated question; if GPT-4o can answer it correctly, we will keep that QA pair in the \papername~Bench. This heuristic filter ensures that the answers are from document figures and double-checks the correctness of generated answers.

\begin{figure}[htp]
    \centering
    \vspace{-1em}
    \begin{minipage}[t]{0.47\textwidth}
        \centering
        \includegraphics[width=\textwidth]{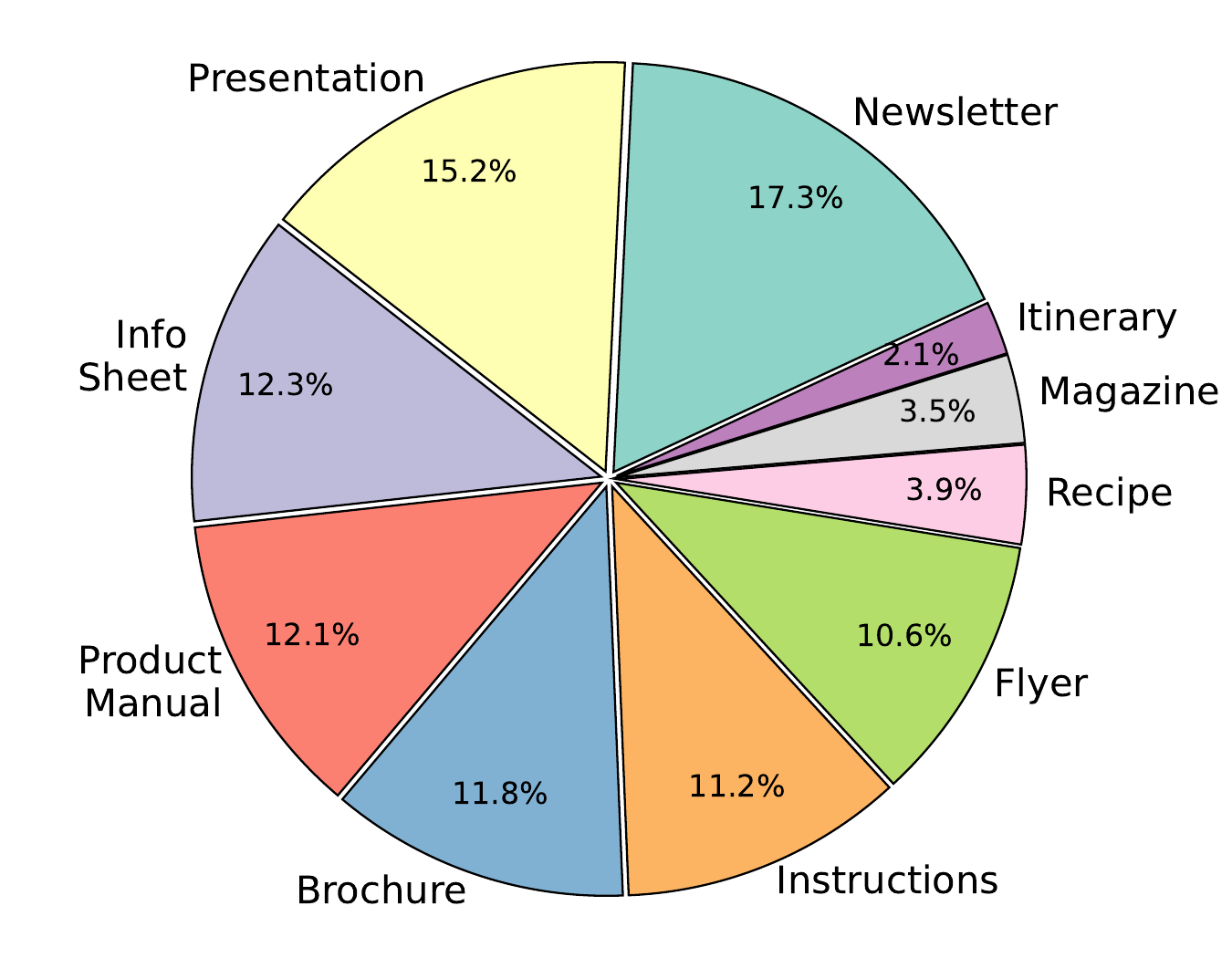}
        \vspace{-1em}
    \end{minipage}
    \hfill
    \begin{minipage}[t]{0.47\textwidth}
        \centering
        \includegraphics[width=\textwidth]{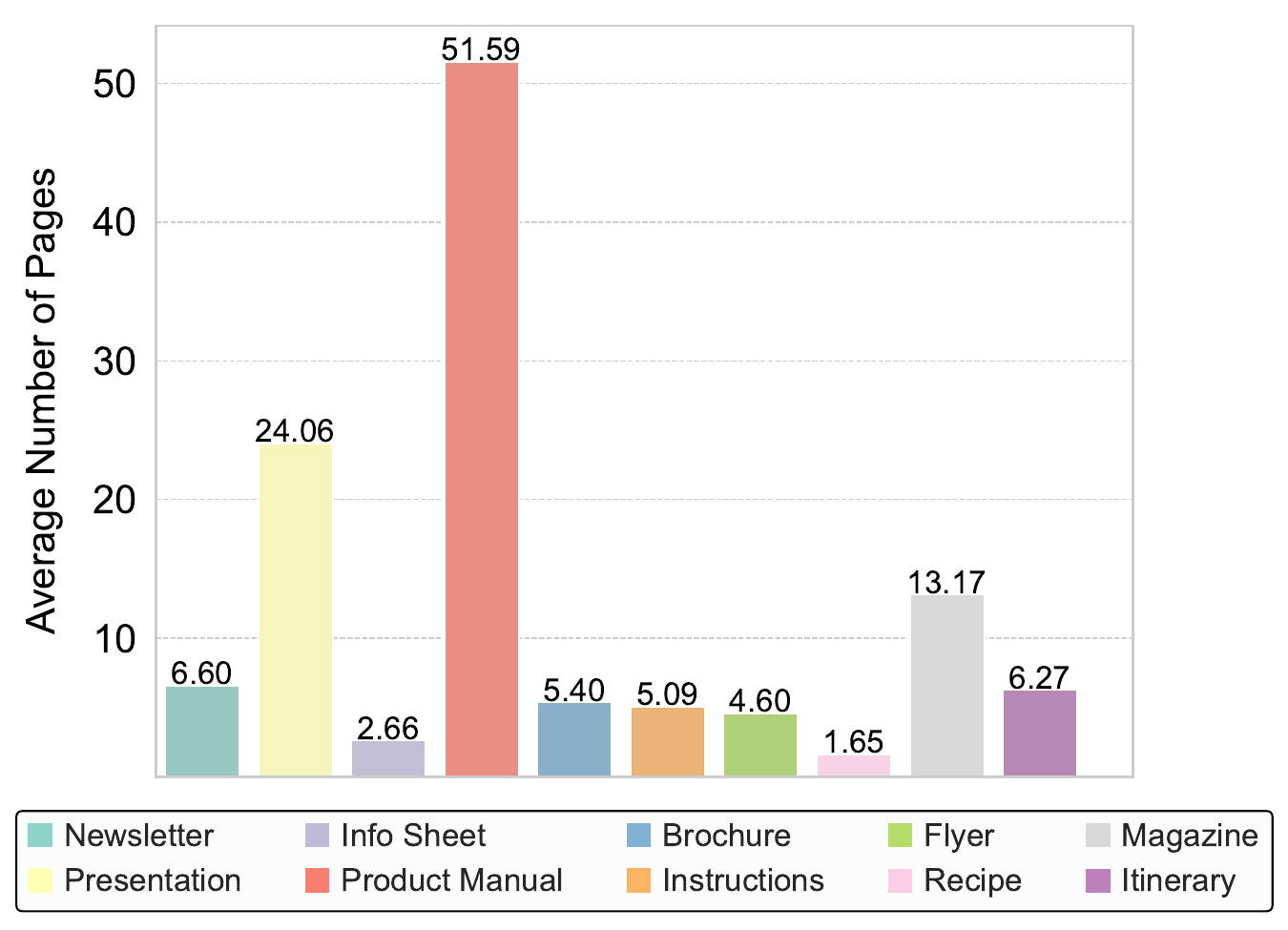}
        \vspace{-1em}
    \end{minipage}
    % \vspace{-1em}
    \caption{Distribution of document types (left) and average document lengths in each types (right).}
    \label{fig:local_stats}
\end{figure}

\paragraph{Dataset Statistics} \dataname-Bench contains 226 documents and 471 human-verified question-answer pairs. Figure \ref{fig:local_stats} shows the distributions of the document types and the length distribution by document type. \dataname-Bench has a great diversity of documents compared to previous work~\citep{tanaka2023slidevqa,islam2023financebench,ma2024mmlongbench}.

\section{Experiments}
We assess the performance of \papername~in evidence page retrieval and visual question answering capabilities.
We first evaluate the retrieval accuracy of the Col-retrieval module within \papername~and compare it with several baselines on SlideVQA~\citep{tanaka2023slidevqa}, MM-LongBench~\citep{ma2024mmlongbench}, DUDE~\citep{van2023document}, DocVQA~\citep{mathew2020docvqa,mathew2021docvqa} and \dataname-Bench. We then conduct experiments on question answering using \papername~and compare the results with other LMM baselines, inlcuding single-page and cross-page VQA. All experiments are implemented with PyTorch and conducted on Nvidia A100 GPUs. The Col-retrieval modules are fine-tuned for 4 epochs with a batch size of 32 and a learning rate of 5e-5, using the AdamW optimizer and LoRA adapters on all linear layers in the LLM. The LoRA rank is set to 32.

\subsection{Datasets}
\paragraph{Finetuning Dataset}
We train our Col-retrieval modules using the original training data of ColPali \citep{faysse2024colpali}, which includes 39,463, 10,074, 13,251, 10,000, and 45,940 question-page pairs filtered from DocVQA, InfoVQA \citep{mathew2022infographicvqa}, TATDQA \citep{zhu2024doc2soargraph}, arXivQA \citep{li2024multimodal}, and synthetic data across various topics, including government reports, healthcare, artificial intelligence, and energy. We incorporated DocMatix-IR \citep{dse} and PFL-DocVQA \citep{tito2023privacy}, using GPT-4o to filter out duplicate images and unsuitable questions. The expanded dataset improved top-1 retrieval accuracy on MMLongBench-Doc by $\sim$1\% without affecting other benchmarks. We fine-tuned our QA modules using the training split of the SlideVQA dataset \citep{tanaka2023slidevqa}. The SlideVQA dataset contains 1,919 slides in the training set, 400 in the test set, and 300 in the development set, with each slide consisting of 20 pages. The training split includes 10,290 samples, each annotated with questions, answers, and corresponding evidence.

\vspace{-1em}
\paragraph{Evaluation Dataset}
We evaluated our method’s performance on four public datasets—SlideVQA, MMLongBench-Doc \citep{ma2024mmlongbench}, DocVQA \citep{mathew2021docvqa}, and DUDE \citep{van2023document}—along with our proposed \dataname-Bench dataset. The evaluation was conducted in both single-evidence (SP) and cross-evidence (MP) settings, where questions require information from either a single page or multiple pages within a long document. For DocVQA, we used 5,349 SP and 5,187 MP QA pairs from the validation split. Similarly, we combined the test and dev splits of SlideVQA to form 2,995 SP and 763 MP QA pairs for evaluation. For DUDE, we evaluated 6,307 QA pairs from the validation split.

MMLongBench-Doc, which consists of 135 PDF documents averaging 50.4 pages (ranging from 9 to 468 pages), contains 1,081 QAs in total. From these, we extracted 488 single-evidence QAs to assess the performance of MLLMs designed for single-image tasks. Additionally, we report the results of our best-performing model across all categories in MMLongBench-Doc, providing a comprehensive comparison against state-of-the-art LMMs.
% \roy{this part need some organizations.}

\subsection{Evaluation Metrics}
\label{sec:metrics}
We evaluate our model’s performance on evidence retrieval and question-answering using several key metrics: Top-k Accuracy, Exact Match (EM) \citep{tanaka2023slidevqa}, Generalized Accuracy (G-Acc) \citep{ma2024mmlongbench}, Average Normalized Levenshtein Similarity (ANLS) \citep{biten2019scene}, and Partial Normalized Levenshtein Similarity (PNLS) \citep{chen2024mmr}. A detailed explanation of each metric can be found in Appendix \ref{sec:appendix_metrics}.

\subsection{Comparative Retrieval Accuracy Analysis}
\label{sec:retreival}
We evaluated the accuracy of the Col-retrieval module in SlideVQA, MMLongBench-Doc, SP-DocVQA, and \dataname-Bench, comparing it with the baseline methods including CLIP (ViT-L/14) \citep{radford2021learning}, SigLip (so400m-patch14-384) \cite{zhai2023sigmoid}, BM25 \citep{robertson2009probabilistic}, SBERT \citep{reimers-2019-sentence-bert}, BGE-M3 \cite{bge-m3}, BGE-large \cite{bge_embedding}, and NV-Embed-v2 \cite{lee2024nv}. For encoder models, we used their text and image encoders to compute the cosine similarity between the feature of the question and the page. For text-based methods, the text content in the MMLongBench-Doc and \papername~Bench datasets is extracted using a document parser to ensure higher accuracy. For SlideVQA and SP-DocVQA, where only scanned images are available, the text is extracted using Paddle-OCR\footnote{PaddleOCR: \textcolor{magenta}{\href{https://github.com/PaddlePaddle/PaddleOCR}{https://github.com/PaddlePaddle/PaddleOCR}}}.

\input{sections/table_retrieval}

The results indicate that Col-retrieval outperforms all baselines, achieving more than 98\% in top-5 retrieval accuracy on the SlideVQA dataset, where each slide consists of 20 pages. However, performance decreases on other datasets as the data become more complex and document lengths increase significantly.

\subsection{Main Results}
We compared the performance of our method with popular lightweight LMMs on document question answering tasks, using PaliGemma \citep{beyer2024paligemma}, Phi-3-v \citep{abdin2024phi}, and InternVL2-4B \citep{chen2023internvl} as the backbone LMMs for both retrieval and QA modules, following the dual adapter design from Section \ref{sec:dual_adapter}. We fine-tuned the retrieval module using the 118,695 training question-page pairs used in ColPali \citep{faysse2024colpali}. The QA module is fine-tuned using SlideVQA’s training split. We reported the original evaluation metrics used in prior works, including EM, G-Acc, and ANLS, and additionally reported PNLS, which better evaluates LLM-generated responses.

Table \ref{tab:sp_results} presents the comparison results. We first evaluate \papername~on single-evidence questions from SP-SlideVQA, MMLongBench-Doc, and SP-DocVQA, where the required information is on a single page. To demonstrate the question-answering capabilities of LMMs, we include four “cheating” baselines where models are given the ground truth evidence page. Next, we test \papername~on cross-evidence questions from MP-SlideVQA, MP-DocVQA, and DUDE, where information spans multiple pages. We only test \papername~with InternVL2-4B backbone, since the other two LMM are pre-trained for single-page understanding. \papername’s performance is compared with classical encoder-only and encoder-decoder models, including BERT \citep{kenton2019bert}, Longformer \citep{beltagy2020longformer}, Big Bird \citep{zaheer2020big}, T5 \citep{raffel2020exploring}, Hi-VT5 \citep{tito2023hierarchical}, and LayoutLMv3 \citep{huang2022layoutlmv3}, with results taken from the best settings in the original papers. InternVL2-8B and GPT-4o, processing all pages, serve as the state-of-the-art baselines for open-source and proprietary multipage LMMs, respectively. We demonstrate how the limitations of the retrieval and QA modules can impact overall performance through challenging examples from the SlideVQA dataset, as shown in Appendix \ref{sec:Failure}. Additional comparisons with text-only baselines utilizing a document parser are provided in Appendix \ref{tab:parser_results}.

\input{sections/table_main}

\paragraph{Retrieval vs Multipage}
We observe \papername~consistently outperforms InternVL2-8B, across various settings. The primary issue with LMMs is that long documents are transformed into excessively long visual token sequences, leading to significant memory burdens, as reported later in section \ref{sec:memory}. In datasets like MMLongBench-Doc and DocVQA, some documents exceed hundreds of pages, causing out-of-memory errors, even on servers with 8 $\times$ A100 (80GB) GPUs. In such cases, we assigned a zero score in our experiments. In contrast, GPT-4o exhibits strong multi-page reasoning capabilities. However, the accuracy of the cheating baseline slightly surpasses that of using all pages, as providing only the evidence pages helps GPT-4o avoid distractions from irrelevant information in the longer context. Moreover, \papername~with InternVL2-4B backbone perform slightly better than the one with Phi-3-vision backbone on MMLongBench-Doc and SP-DocVQA, possibly due to the improvement in retrieval accuracy by using top-5 pages, which is more crucial for longer documents. 

\paragraph{Impact of Fine-tuning}
We observe that \papername~QA modules with PaliGemma and InternVL2-4B backbones show a significant increase in EM on the SlideVQA dataset, surpassing their cheating baselines after fine-tuning on SlideVQA. The model with the Phi-3-vision backbone shows notable improvements in Exact Match (EM) scores without gains in PNLS, suggesting that fine-tuning primarily enhanced the model’s attention and answer formatting. This could be because the pre-trained model was already optimized for these question types. Nevertheless, as shown in Figure \ref{fig:QA_demo}, we empirically find that fine-tuning still improves answering performance. However, we notice a performance drop for fine-tuned \papername-Phi-3-vision on MMLongBench-Doc, indicating that fine-tuning can harm LLM generalization. A similar trend is seen with the InternVL2-4B backbone on the DUDE dataset. 

\paragraph{Comparison with SOTA LMMs}
Finally, we present the complete results of \papername-InternVL2-4B on the MMLongBench-Doc dataset to highlight the advantages of our method. As shown in Table \ref{tab:MMLongBench}, our model, with only 4 billion parameters, outperforms all open-source LMMs and achieves performance comparable to proprietary models such as Claude-3 Opus and Gemini-1.5-Pro.
\input{sections/table_MMLong}

\vspace{-1em}
\subsection{Efficiency of Different Models}
\label{sec:memory}
To evaluate the efficiency of \papername, we conducted experiments on the SlideVQA dataset, which has 20 pages per question with a resolution of 1024x768. We recorded peak GPU memory usage and time costs for retrieval and QA modules separately. The GPU memory is manually recorded using the \texttt{nvidia-smi} command, which tends to report higher numbers than the actual memory required by the application due to overhead and memory management processes. We tested backbones including PaliGemma, Phi-3-v, and InternVL2-4B, all equipped with LoRA adapters. Since PaliGemma and Phi-3-v are single-page models, we used top-1 retrieved image as input. InternVL2-4B, however, supports multi-image input, allowing us to test with the top-1, 5, and 12 retrieved images.

\input{sections/table_efficiency}
As shown in Table \ref{tab:cost}, the QA module’s memory consumption increases with the number of evidence pages, with 13 images (1024x768) exceeding the 80GB limit on an A100 GPU resulting in out-of-memory error. In contrast, the retrieval module maintains low memory usage, as \papername~processes pages independently, with costs equivalent to single-page reasoning. Although multi-evidence QA requires more memory, \papername~remains efficient and compact, making it well-suited for answering questions from fewer evidence pages in resource-constrained environments. This demonstrates \papername’s ability to balance performance and resource usage, ensuring scalability across diverse deployment scenarios. Additional results on the retrieval efficiency are presented in Table \ref{tab:retrieval_time}.

\subsection{Ablation}
\label{sec:ablation}
In our experiment, we use the hidden states from the last transformer layer (index 31) as the feature sequence. However, LLMs consist of multiple transformer layers, each encoding different types of information. To assess the impact of layer selection, we conduct an ablation study on the hidden states used to compute the late interaction score in Eq.(\ref{eq:late_interaction}). Given the high computational cost of training the col-retrieval module across all layers, we instead evaluate top-1 accuracy on the MMLongBench-Doc dataset using hidden states from different layers of the Phi-3-vision model with pre-trained weights.
\begin{figure}[!h]
    \centering
    \includegraphics[width=0.65\textwidth]{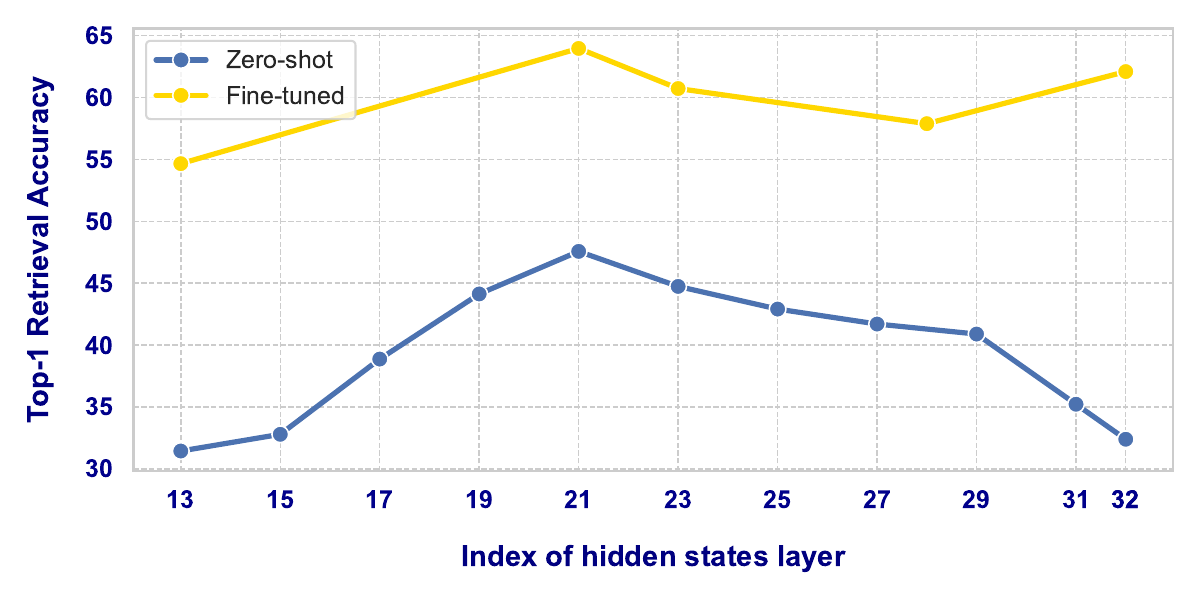}
    \vspace{-1em}
    \caption{Top-1 retrieval accuracy on MMLongBench-Doc using different hidden states across all layers of Phi-3-vision.}
    \label{fig:ablation_curve}
    \vspace{-1em}
\end{figure}

Figure \ref{fig:ablation_curve} shows that the hidden states of the 21th layer yield the highest accuracy. After fine-tuning a model with hidden states from this layer, we observed improved accuracy compared to using hidden states of the final layer. In particular, using hidden states from earlier layers can significantly reduce computational costs, enabling faster retrieval during inference. 

% However, since the accuracy gain is marginal, we chose to use the hidden states of the final layer in our main experiments for consistency. Further exploration of this optimization is left for future work. 

%% file: sections/table_retrieval.tex
\begin{table}[!h]
\centering
\fontsize{9}{11}\selectfont
\setlength\arrayrulewidth{0.6pt}
\resizebox{0.9\textwidth}{!}{
\begin{tabular}{llcclcclcclcc}
\hline
\multicolumn{1}{r}{} & &  \multicolumn{2}{c}{\textbf{SlideVQA}} &  & \multicolumn{2}{c}{\textbf{MMLong}} & &  \multicolumn{2}{c}{\textbf{\dataname-B}} & & \multicolumn{2}{c}{\textbf{SP-DocVQA}} \\ 
 \cline{3-4}  \cline{6-7}  \cline{9-10}  \cline{12-13}
\multicolumn{1}{r}{accuracy} & &  top1 & top5  & & top1 & top5 &  & top1 & top5 &  & top1 & top5 \\ \hline
\multicolumn{2}{l}{\textcolor{gray}{\fontsize{7}{8}\textit{Text-based Methods}}} &&&&&&&&&&& \\
BM25 &  & 69.3 & 91.1 & &  25.3 & 47.6  & & 32.2 & 57.5 & & 30.9 & 61.7 \\ 
SBERT &  & 73.0 & 91.0 & &  44.7 & 70.2 & & 38.8 & 72.1 & & 47.4 & 74.0 \\ 
BGE-M3 & & 74.3 & 92.0 & & 42.7 & 66.6 & & 47.7 & 78.1 & & 47.8 & 77.5\\ 
Bge-large & & 81.3 & 93.3 & & 47.4 & 71.5 & & 53.7 & 80.3 & & 56.7 & 81.5\\ 
NV-Embed-v2 & & 82.2 & 94.3 & & 47.4 & 69.0 & & 55.2 & 82.7 & & 51.7 & 80.2\\ 
\hline 
\multicolumn{2}{l}{\textcolor{gray}{\fontsize{7}{8}\textit{Encoder Models}}} &&&&&&&&&&& \\
CLIP & & 58.4 & 86.9 & &  32.4 & 63.4 & & 33.4 & 62.1 & &  37.1 & 69.4\\ 
SigLip & & 66.2 & 90.1 & & 44.9 & 69.4 & & 53.2 & 81.3 & & 39.3 & 71.9\\ 
\hline 
\rowcolor{SVRAG} \multicolumn{2}{l}{\textcolor{gray}{\fontsize{7}{8}\textit{Col-Retrieval Modules}}} &&&&&&&&&&& \\
\rowcolor{SVRAG} Col-Paligemma &  & 89.0 & 98.7  & & 60.7 & 82.0 & &  67.9 & 90.8  & & 62.3 & 85.9 \\  
\rowcolor{SVRAG} Col-InternVL2 &  & 88.5 & 98.3 &  & 61.3 & 83.0 &  & 69.3 & 90.7 & &  63.2 & 85.9 \\ 
\rowcolor{SVRAG} Col-Phi-3-vision &  & {90.6} & {98.8} &  & {64.8} & {84.8} &  & {71.9} & {91.8}  & & {65.1} & {87.0} \\ 
\hline
\end{tabular}
}
\caption{Retrieval accuracy results on four datasets, where MMLong refers to MMLongBench-Doc, \papername-B refers to \papername-Bench. Bold font indicates the best model.}
\label{tab:retrieval}
\end{table}

%% file: sections/table_main.tex
\begin{table}[!h]
\centering
\caption{\textbf{Quantitative Results in Multi-Page QA}: "\#Param" refers to number of parameters. "Evidence" reports evidence setting: T (true evidence page), A (all pages), and Rk (top-k retrieved). Reported metrics include PNLS, Exact Match, Generalized Accuracy, and ANLS. $\dagger$ indicates models with LoRA adapter on QA module. Results for all encoder/decoder models are taken from their respective papers, with “-” indicating missing or not applicable results. Bold font indicates the best open-source model, excluding cheating baselines.}
\label{tab:sp_results}
\fontsize{9}{11}\selectfont
\setlength\arrayrulewidth{0.6pt}
\resizebox{1\textwidth}{!}{
\begin{tabular}{lcclcclcclcc}
\hline
\multirow{2}{*}{Method} & \multirow{2}{*}{\#Param} & \multirow{2}{*}{Evidence} & &  \multicolumn{2}{c}{\textbf{SP-SlideVQA}} & & \multicolumn{2}{c}{\textbf{MMLongBench}}  & & \multicolumn{2}{c}{\textbf{SP-DocVQA}} \\ 
\cline{5-6} \cline{8-9} \cline{11-12}
 &  &   & & EM & PNLS & &  G-Acc & PNLS & &  ANLS & PNLS  \\ \hline
\rowcolor{LLightGray} \multicolumn{12}{c}{\textit{Single-Page Evidence}} \\
\hline
\textcolor{gray}{\fontsize{7}{8}\textit{Cheating Baselines}} &&&&&& \\ 
PaliGemma  & 3B & T  & & 37.30 & 0.63  & & 23.9 & 0.38  & & 0.65 & 0.79 \\
Phi-3-v &  4B & T  & & 13.72 & 0.80  & & 33.7 & 0.52 & &  0.65 & 0.85 \\
InternVL2 & 4B & T  & & 15.03 & 0.58 & &  40.4 & 0.55 & &  0.84 & 0.88 \\
GPT-4o & - & T  & & 30.59 & 0.84 &  & 56.8 & 0.62 & &  0.87 & 0.94 \\ \hline
\textcolor{gray}{\fontsize{7}{8}\textit{Multi-image MLLMs}} &&&&&& \\
InternVL2 & 8B & A  & & 12.62 & 0.65  & & 14.1 & 0.22  & & 0.50 & 0.55 \\
GPT-4o & - & A  & & 27.28 & 0.81 & &  54.5 & 0.57 &  & 0.69 & 0.80 \\ \hline 
\rowcolor{SVRAG} \textcolor{gray}{\fontsize{7}{8}\textit{\papername~Models (Proposed)}} &&&&&&&&&&& \\
\rowcolor{SVRAG} \papername-PaliGemma &  3B & R1  & & 35.03 & 0.60 & &  23.9 & 0.35 &  & 0.56 & 0.69 \\ 
\rowcolor{SVRAG} \papername-PaliGemma$^{\dagger}$ & 3B & R1 & &  49.75 & 0.65  & & 23.1 & 0.38 & &  0.56 & 0.68 \\ 
\rowcolor{SVRAG} \papername-Phi-3-vision &  4B & R1  & & 12.85 & \textbf{0.78}  & & 30.7 & \textbf{0.50} &  & 0.55 & 0.75 \\
\rowcolor{SVRAG} \papername-Phi-3-vision$^{\dagger}$ & 4B & R1  & & \textbf{58.13} & 0.77 &  & 28.4 & 0.44 & &  0.68 & 0.73 \\
\rowcolor{SVRAG} \papername-InternVL2 & 4B & R5  & & 16.40 & 0.58  & & 33.2 & 0.48  & & 0.70 & \textbf{0.76} \\ 
\rowcolor{SVRAG} \papername-InternVL2$^{\dagger}$ &  4B & R5 &  & 45.07 & 0.77 & &  \textbf{34.0} & 0.49 & &  \textbf{0.71} & 0.75 \\ \hline \hline
% Next page
\rowcolor{LLightGray} \multicolumn{12}{c}{\textit{Cross-Page Evidence}} \\
\hline
\multirow{2}{*}{Method} & \multirow{2}{*}{\#Param} & \multirow{2}{*}{Evidence} &  &  \multicolumn{2}{c}{\textbf{MP-SlideVQA}}  &  & \multicolumn{2}{c}{\textbf{MP-DocVQA}}  & & \multicolumn{2}{c}{\textbf{DUDE}} \\ 
\cline{5-6} \cline{8-9} \cline{11-12}
 &  &  & &  EM & PNLS &  &  ANLS & PNLS &  & ANLS & PNLS \\ \hline
\textcolor{gray}{\fontsize{7}{8}\textit{Encoder/Decoder models}} &&&&&& \\ 
BERT-Large & 334M & - & & - & - &  & 0.53 & -  & & 0.25 & - \\
Longformer & 148M & - & &  - & - &  & 0.55 & - & &  0.27 & - \\
Big Bird & 131M & - &  & - & - &  & 0.58 & - &  & 0.26 & - \\
T5-Base & 223M & - &  & -  & - &  & 0.51 & - & &  0.42 & - \\
LayoutLMv3 & 125M & - & &  - & - &  & 0.55 & - & &  0.20 & - \\
Hi-VT5 & 316M & - &  & - & - &  & 0.62 & - &  & 0.23 & - \\ 
% T5-2D-Large & 770M & - &  & - & - &  & - & - & &  0.46 & - \\ 
\hline
\textcolor{gray}{\fontsize{7}{8}\textit{Multi-image MLLMs}} &&&&&& \\
InternVL2 & 8B & A &  & 17.04 & 0.53 & &  0.68 & 0.75 &  & 0.37 & 0.56 \\
GPT-4o & - & A &  & 16.09 & 0.73 & &  0.67 & 0.79 &  & 0.54 & 0.70 \\ \hline 
\rowcolor{SVRAG} \textcolor{gray}{\fontsize{7}{8}\textit{\papername~Models (Proposed)}} &&&&&&&&&&& \\
% \papername-PaliGemma &  3B & R1 &  &  &  &  &  &  &  &  \\ 
% \papername-PaliGemma$^{\dagger}$ & 3B & R1 & &  &  &    \\ 
% \papername-Phi-3-vision &  4B & R1 & &  &  &  &  &  &  &  \\
% \papername-Phi-3-vision$^{\dagger}$ & 4B & R1 & &  &  &  &  &  &  \\
\rowcolor{SVRAG} \papername-InternVL2 &  4B & R5 & &  24.25 & \textbf{0.61} & &  0.70 & \textbf{0.76} &  & 0.36 & \textbf{0.57} \\ 
\rowcolor{SVRAG} \papername-InternVL2$^{\dagger}$ & 4B & R5 & &  \textbf{31.98} & 0.59 & &  \textbf{0.71} & \textbf{0.76} & &  \textbf{0.45} & 0.54  \\ \hline 
\end{tabular}
}
\end{table}

%% file: sections/table_MMLong.tex
\begin{table}[!ht]
\centering
\fontsize{9}{12}\selectfont
\setlength\arrayrulewidth{0.6pt}
\resizebox{1\textwidth}{!}{
\begin{tabular}{lc|ccccc|ccc|cc}
\hline
 \multirow{2}{*}{Method} & \multirow{2}{*}{\#Param} & \multicolumn{5}{c|}{\textbf{Evidence Source}} & \multicolumn{3}{c|}{\textbf{Evidence Page}} &  \multirow{2}{*}{ACC} & \multirow{2}{*}{F1} \\
 & & TXT & LAY & CHA & TAB & FIG & SIN & MUL & UNA &  & \\
 \hline
\textcolor{gray}{\fontsize{7}{8}\textit{Open-source Models}} &&&&&&&&&\\
DeepSeek-VL-Chat & 7.3B & 7.2 & 6.5 & 1.6 & 5.2 & 7.6 & 5.2 & 7.0 & 12.8 & 7.4 & 5.4 \\
Idefics2 & 8B  & 9.0 & 10.6 & 4.8 & 4.1 & 8.7 & 7.7 & 7.2 & 5.0 & 7.0 & 6.8 \\
MiniCPM-Llama3-V2.5 & 8B & 11.9 & 10.8 & 5.1 & 5.9 & 12.2 & 9.5 & 9.5 & 4.5 & 8.5 & 8.6 \\
InternLM-XC2-4KHD & 8B  & 9.9 & 14.3 & 7.7 & 6.3 & 13.0 & 12.6 & 7.6 & 9.6 & 10.3 & 9.8 \\
mPLUG-DocOwl 1.5 & 8.1B & 8.2 & 8.4 & 2.0 & 3.4 & 9.9 & 7.4 & 6.4 & 6.2 & 6.9 & 6.3 \\
Qwen-VL-Chat & 9.6B  & 5.5 & 9.0 & 5.4 & 2.2 & 6.9 & 5.2 & 7.1 & 6.2 & 6.1 & 5.4 \\
Monkey-Chat & 9.8B & 6.8 & 7.2 & 3.6 & 6.7 & 9.4 & 6.6 & 6.2 & 6.2 & 6.2 & 5.6 \\
CogVLM2-LLaMA3-Chat & 19B  & 3.7 & 2.7 & 6.0 & 3.2 & 6.9 & 3.9 & 5.3 & 3.7 & 4.4 & 4.0 \\
InternVL-Chat-v1.5 & 26B & 14.0 & 16.2 & 7.1 & 10.1 & 16.6 & 14.9 & 12.2 & \textbf{17.5} & 14.6 & 13.0 \\
EMU2-Chat & 37B & 6.1 & 9.7 & 2.6 & 3.8 & 7.7 & 5.7 & 6.1 & 16.5 & 8.3 & 5.5 \\ \hline
\rowcolor{SVRAG}  \textcolor{gray}{\fontsize{7}{8}\textit{\papername~Models (Proposed)}} &&&&&&&&&&&\\
\rowcolor{SVRAG} \papername-InternVL2 (R5) & 4B   & \textbf{26.5} & 18.8 & 22.3 & 19.6 & 23.6 & 33.2 & \textbf{13.1} & 12.4 & 22.2 & 22.8 \\
\rowcolor{SVRAG} \papername-InternVL2$^{\dagger}$ (R5) & 4B   & 26.3 & \textbf{22.1} & \textbf{25.0} & \textbf{20.7} & \textbf{25.2} & \textbf{34.0} & 10.6 & 15.7 & \textbf{23.0} & \textbf{24.2} \\ \hline
\textcolor{gray}{\fontsize{7}{8}\textit{Proprietary Models}} &&&&&&&&&&&\\
Claude-3 Opus & -   & 24.9 & 24.7 & 14.8 & 13.0 & 17.1 & 25.6 & 13.8 & 7.6 & 17.4 & 18.1 \\
Gemini-1.5-Pro & -   & 21.0 & 17.6 & 6.9 & 14.5 & 15.2 & 21.1 & 11.1 & 69.2 & 28.2 & 20.6 \\
GPT-4V & -   & 34.4 & 28.3 & 28.2 & 32.4 & 26.8 & 36.4 & 27.0 & 31.2 & 32.4 & 31.2 \\
GPT-4o & -   & 46.3 & 46.0 & 45.3 & 50.0 & 44.1 & 54.5 & 41.5 & 20.2 & 42.8 & 44.9 \\ \hline 
\end{tabular}
}
\caption{\textbf{Performance of various models on MMLongBench-Doc.} Questions are categorized in two ways: (1) by evidence source type—text (TXT), layout (LAY), chart (CHA), table (TAB), and image (IMG); and (2) by evidence pages—single-page (SIN), cross-page (MUL), and unanswerable (UNA). Models using LoRA adapters fine-tuned on SlideVQA for the QA module are marked with $\dagger$. Bold font indicates the best open-source model. The results of baseline models are adopted from the original MMLongBench-Doc paper \cite{ma2024mmlongbench}.}
\label{tab:MMLongBench}
\end{table}

%% file: sections/table_efficiency.tex
\begin{table}[!ht]
\centering
\fontsize{9}{11.5}\selectfont
\setlength\arrayrulewidth{0.6pt}
\resizebox{0.65\textwidth}{!}{
\begin{tabular}{lclcclcc}
\hline
 \multirow{2}{*}{\papername-Backbone} & \multirow{2}{*}{Page} & &  \multicolumn{2}{c}{\textbf{Retrieval}}  & & \multicolumn{2}{c}{\textbf{QA}}  \\ 
 \cline{4-5}  \cline{7-8}
 & &  & Time & Mem & &  Time & Mem  \\ \hline
Paligemma & R1 &  &  2.3 & 9.2 &  & 1.0 & 12.4 \\
Phi-3-vision & R1 &  & 4.1 & 11.6 &  & 0.9 & 12.9 \\ \hline 
InternVL2-4B & R1  & & 9.2 & 14.2 &  & 1.4 & 14.6 \\
InternVL2-4B & R5  & & 9.2 & 14.2 &  & 2.8 & 40.8 \\
InternVL2-4B & R12  & & 9.2 & 14.2 &  & 4.1 & 76.4 \\
\hline
\end{tabular}
}
\caption{Time (s) cost and Peak GPU memory (GB) cost of \papername~models with different backbones.}
\label{tab:cost}
\end{table}

%% file: sections/appendix.tex
\newpage
\appendix
\counterwithin{figure}{section}
\counterwithin{equation}{section}
\counterwithin{table}{section}

\section{Example of training pairs for retrieval module}
\begin{figure}[!h]
    \centering
    \vspace{-1em}
    \includegraphics[width=0.9\textwidth]{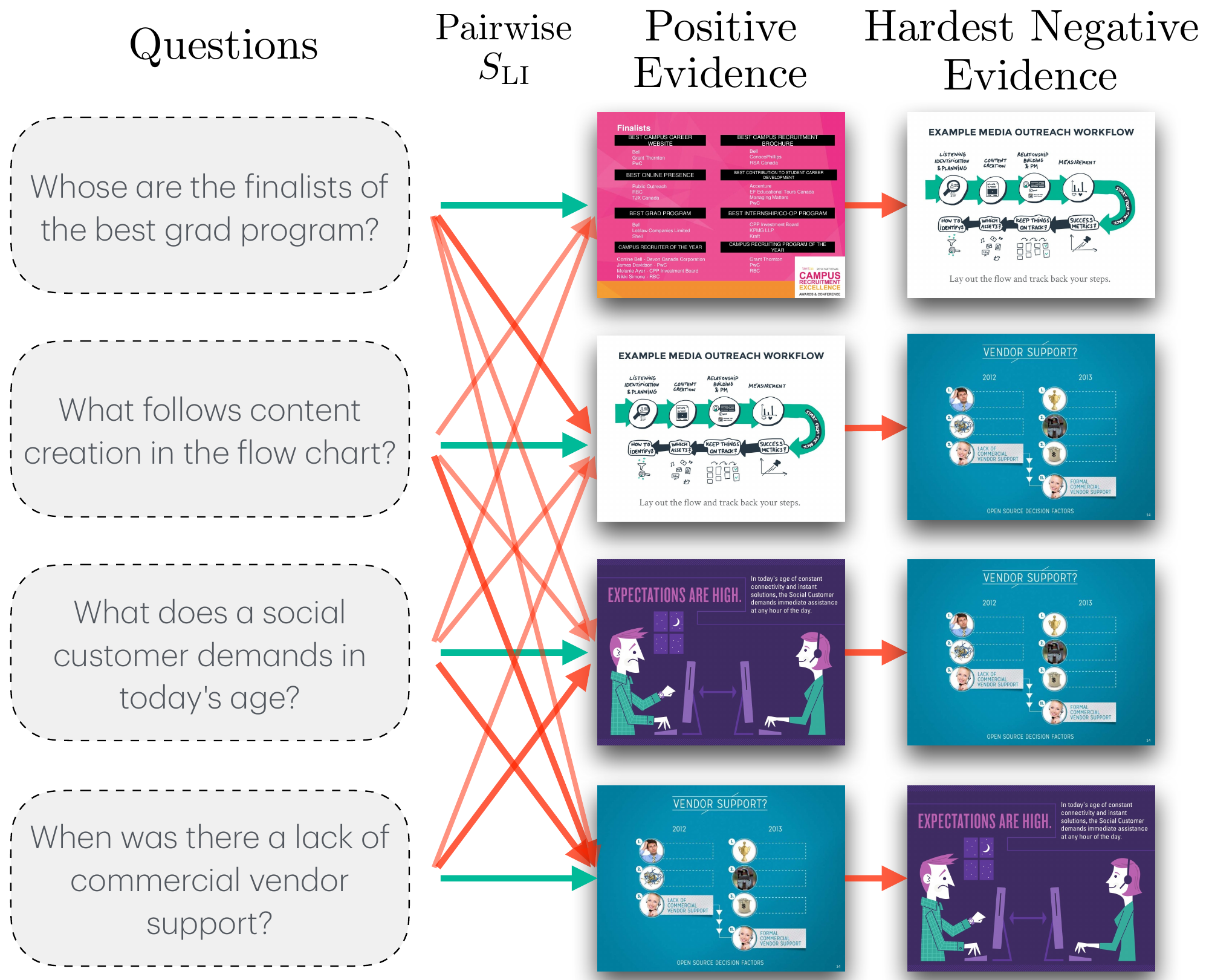}
    \caption{Example of training pairs within a batch (batch size: 4) for contrastive training, using samples from the SlideVQA dataset.}
    \label{fig:negative}
    % \vspace{-1em}
\end{figure}

\section{Evaluation Metrics}
\label{sec:appendix_metrics}
We evaluate the model's performance on evidence retrieval and question-answering using five metrics explained as follows:
\vspace{-0.5em}
\paragraph{Top-k Accuracy}
In our experiment, we focus on questions that have evidence from a single page. We use top-k accuracy to evaluate retrieval methods, which measures the percentage of times the evidence image appears within the top k most similar images. 
\vspace{-0.5em}
\paragraph{Exact Match}
Following \citep{tanaka2023slidevqa}, we report exact match (EM) frequency between generated answers and the ground truth, allowing for case insensitivity and extra spaces. While effective for fine-tuned models, this metric is less suited for LLM responses, which often include full sentences. Correct answers with extra context may thus be unfairly penalized.
\vspace{-0.5em}
\paragraph{Generalized Accuracy}
We report generalized accuracy (G-Acc) from MMLongBench-Doc \citep{ma2024mmlongbench}, a GPT-dependent, rule-based evaluation protocol . Model responses are simplified using GPT-4o and scored based on answer-type-specific rules. However, G-Acc has two limitations: it introduces randomness from GPT’s stochastic outputs and relies on answer-type annotations, limiting its applicability across datasets.
\vspace{-0.5em}
\paragraph{ANLS}
Average Normalized Levenshtein Similarity (ANLS) \citep{biten2019scene} measures the similarity between predicted and ground truth text using the Levenshtein distance, normalized by the longer string’s length. It outputs a similarity score between 0 and 1. ANLS allows mismatches, insertions, and deletions making it useful for OCR and document understanding tasks when exact matches are not required.
\vspace{-0.5em}
\paragraph{PNLS}
The \textit{partial normalized Levenshtein similarity} (PNLS) \citep{chen2024mmr} generalizes ANLS by not penalizing extra prefixes or suffixes while allowing mismatches, insertions, and deletions within the matched region. This makes it more suitable for evaluating LLM responses, which are often verbose to improve user experience. 

The PNLS metric is formally defined as follows: String $\mathcal{T}_{1,m}=t_1\dots t_m$ represents the true answer and $\mathcal{S}_{1,n}=s_1\dots s_n$ is a model generated string. We first use  using the approximate string matching algorithm \citep{sellers1980theory} to identify the sub-string of $\mathcal{S}$ that has the minimum edit distance to $\mathcal{T}$. Specifically, we first construct a scoring matrix $\mathbf{F}$ of size $(m+1)\times(n+1)$, where $F_{i,j}$ stores the smallest edit distance between the $i$-prefix ${\mathcal T}_{1,i}$ and any sub-string ${\mathcal S}_{x, j}$, $\forall x \in \{1,\dots,j-1\}$ that ends at position $j$. The scoring matrix can be computed recursively
\begin{equation*}
\label{recur}
\small
    F_{i,j} = \left\{
                \begin{array}{ll}
                  0 &\text{if } i = 0 \\
                  m &\text{if } j = 0\\
                  \min \left (
                \begin{array}{l}
                F_{i-1,j-1} + c(t_i,s_j)\\
                F_{i-1,j} + 1\\
                F_{i,j-1} + 1 
                \end{array}
                \right )
                &\text{otherwise},\\
                \end{array}
              \right.
\end{equation*}
where $c$ is the substitution cost that takes a value of $0$ if $t_i = s_j$ and $1$ otherwise. Once ${\bf F}$ is computed, the minimum value in the last row is the optimal edit distance and the end index of the matched sub-string $j'=\argmin_{j}(F_{m+1,j})$. The start index $i'$ can be found by tracing back the the computation of Eq.(\ref{recur}) using $\argmin$ operation. Finally, the PNLS is computed as: $m / (m + j'-i'+1)$. In our experiments we use binary cost function: $c(t_i,s_j)=0$ if $t_i=s_j$ else $c(t_i,s_j)=1$

\section{Example of Inference Failure Scenario}
\label{sec:Failure}
\begin{figure}[!h]
    \centering
    \vspace{-1em}
    \includegraphics[width=1\textwidth]{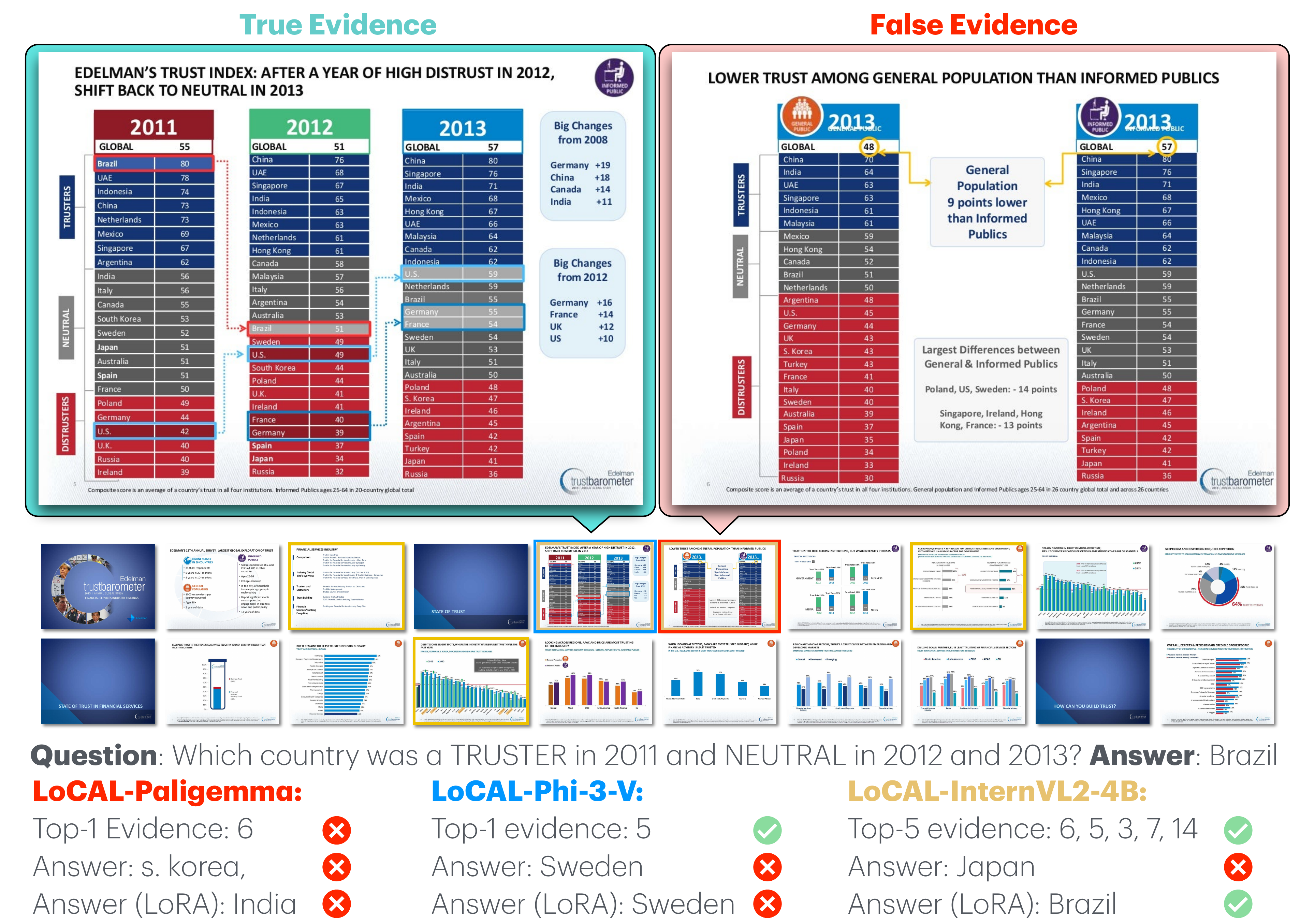}
    \caption{Inference example of a challenging case in the SlideVQA dataset. \papername-Paligemma retrieved the wrong evidence page due to limitations in its retrieval module, leading to an incorrect answer. \papername-Phi-3-V retrieved the correct page but provided a wrong answer due to limitations in its QA module. Meanwhile, \papername-InternVL2-4B also assigned the highest relevance score to an incorrect page. However, since it processes multiple pages (top 5), the correct evidence page was included in the input, allowing its fine-tuned QA module to deliver the correct answer.}
    \label{fig:failure}
    % \vspace{-1em}
\end{figure}

\clearpage
\subsection{Additional Examples of Retrieval Failures}
\begin{figure}[!h]
    \centering
    % \vspace{-1em}
    \includegraphics[width=1\textwidth]{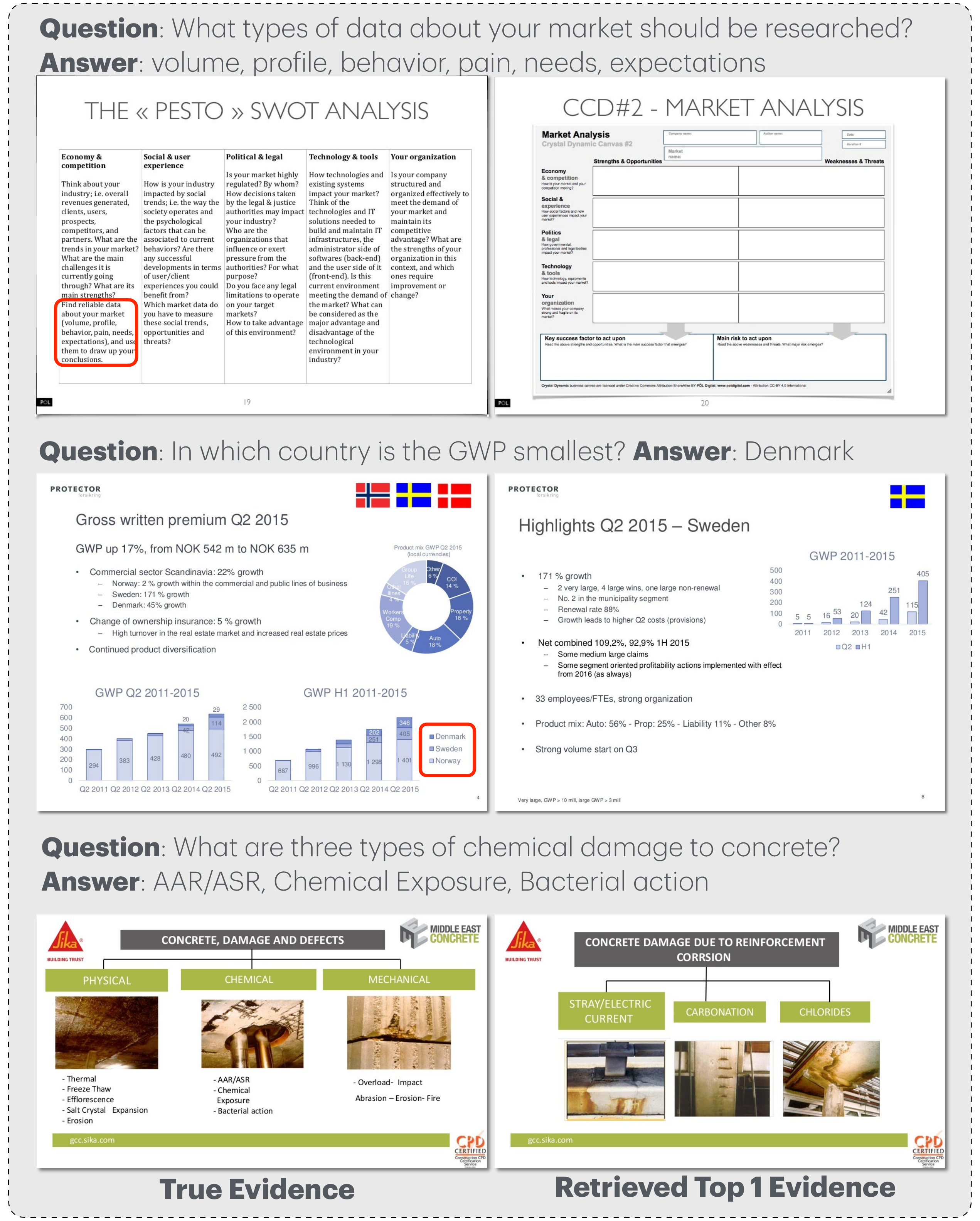}
    \caption{Failure cases from the SlideVQA dataset, highlighting retrieval module errors. In the first two examples, some of the relevant information (highlighted in red boxes) on the true evidence pages is difficult even for human eyes to detect. In the third example, the retrieved page has a high similarity to the true evidence page, making it challenging to rank correctly. Additionally, answering the question accurately requires a deep understanding of the concept of chemical damage and related topics.}
    \label{fig:retrieval_fail}
    % \vspace{-1em}
\end{figure}

\clearpage
\section{Qaulitative Results in Question-Answering}
\begin{figure}[!h]
    \centering
    \vspace{-1em}
    \includegraphics[width=1\textwidth]{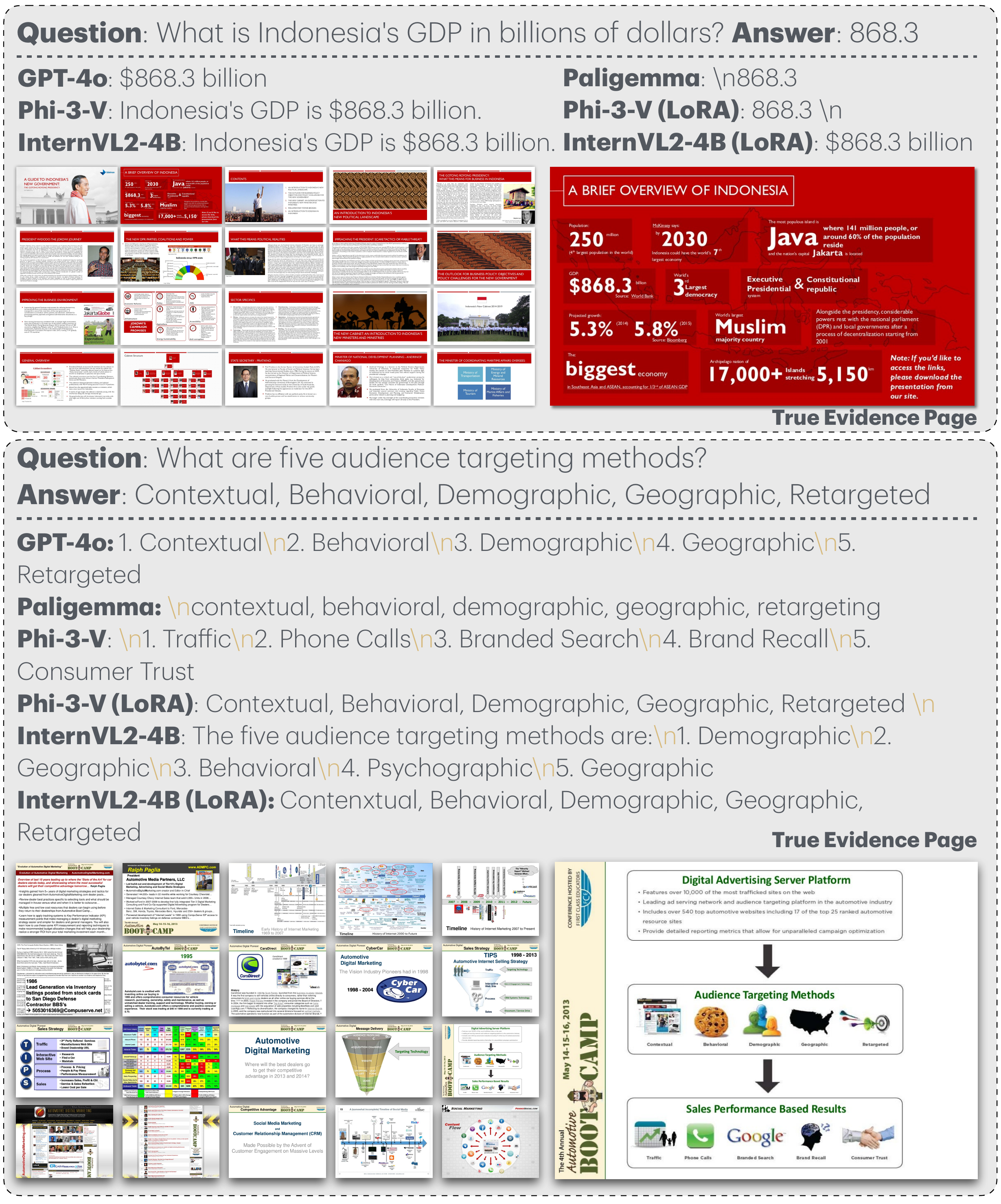}
    \caption{Question answering examples on the SlideVQA dataset using different QA modules. Models without fine-tuning, such as Phi-3-V and InternVL2-4B, tend to produce verbose and error-prone responses. However, in the second example, fine-tuning with the LoRA adapter significantly improves the accuracy of Phi-3-V and InternVL2-4B.}
    \label{fig:QA_demo}
    % \vspace{-1em}
\end{figure}

\clearpage
\section{Examples from the \dataname-Bench dataset}
\begin{figure}[!h]
    \centering
    \vspace{-1em}
    \includegraphics[width=1\textwidth]{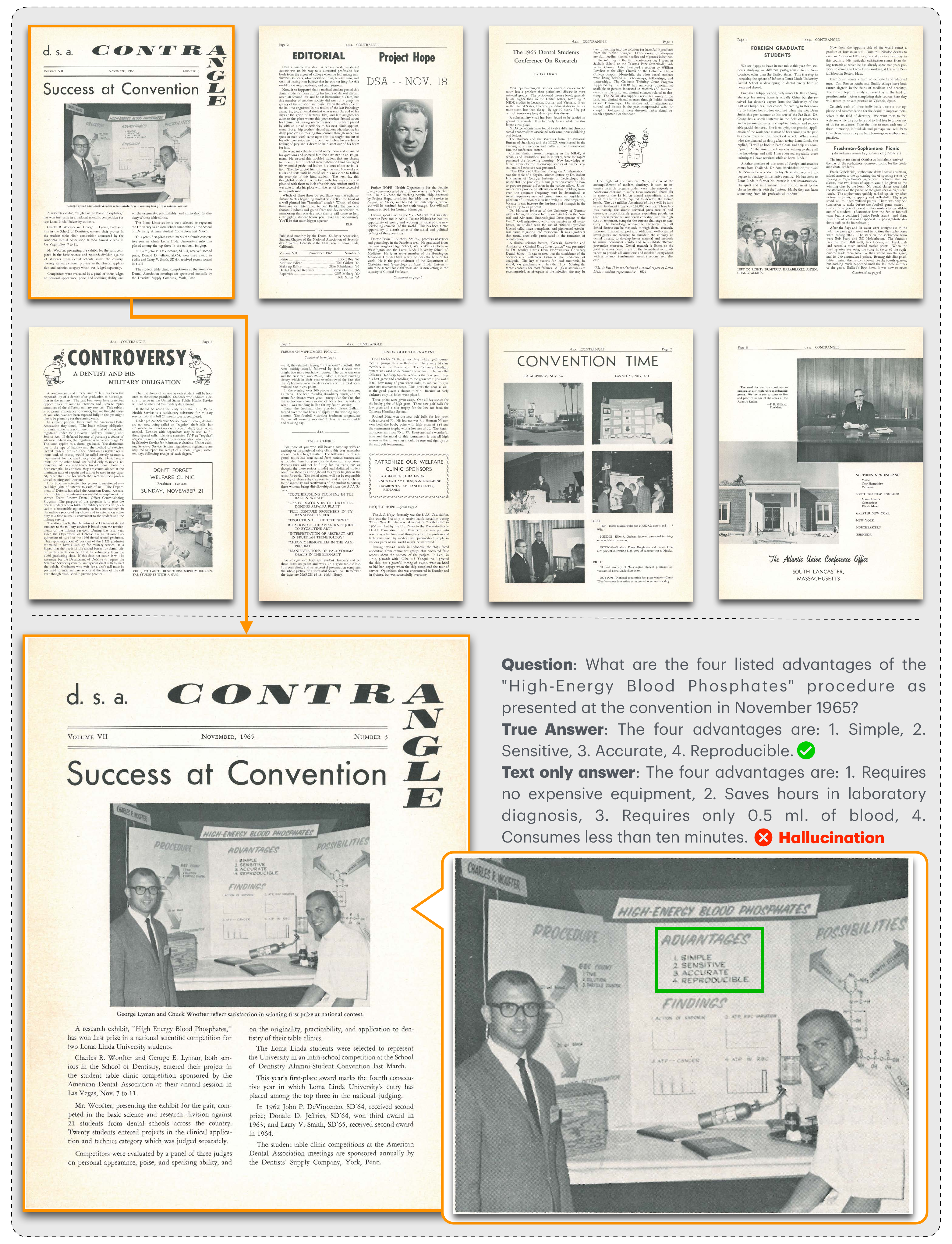}
    \caption{Example question-and-answer pair from the \dataname-Bench dataset, highlighting the reliance on both image and surrounding text for accurate responses.}
    \label{fig:SVRAG_Bench_demo_1}
    % \vspace{-1em}
\end{figure}

\begin{figure}[!h]
    \centering
    \vspace{-1em}
    \includegraphics[width=1\textwidth]{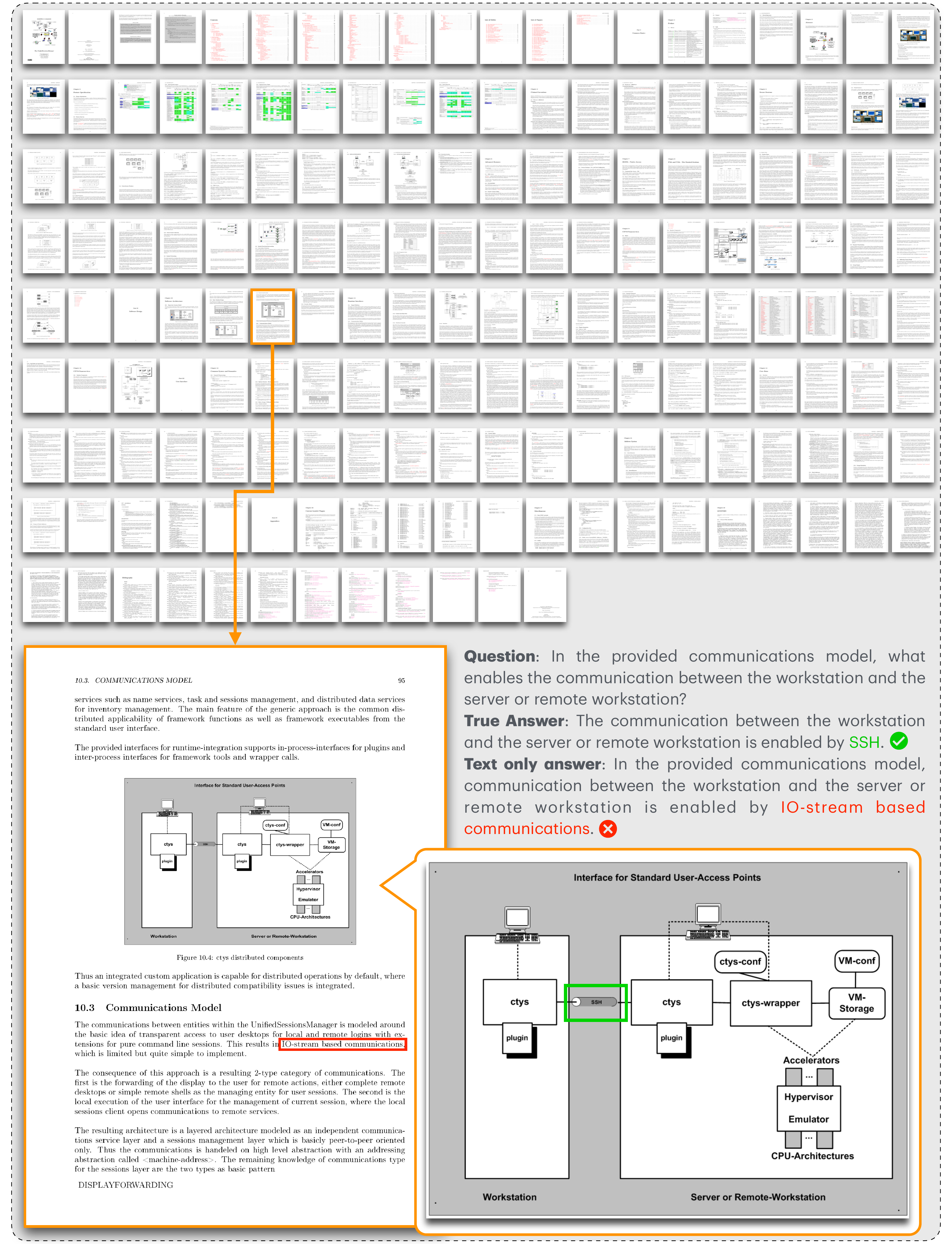}
    \caption{Example question-and-answer pair from the \dataname-Bench dataset, highlighting the reliance on both image and surrounding text for accurate responses.}
    \label{fig:SVRAG_Bench_demo_2}
    % \vspace{-1em}
\end{figure}

\clearpage
\section{Comparison of Retrieval Method Efficiency}
\begin{table}[!h]
\centering
\fontsize{9}{11}\selectfont
\setlength\arrayrulewidth{0.6pt}
\resizebox{0.9\textwidth}{!}{
\begin{tabular}{lllllll}
\cline{1-4} \cline{6-7}
\multicolumn{2}{l|}{Text Extraction} & \multicolumn{2}{l}{Text Encoding} & \hspace{2em} & \multicolumn{2}{l}{Multimodal Encoding} \\ \cline{1-4} \cline{6-7} 
\multirow{2}{*}{PaddleOCR} & \multicolumn{1}{l|}{\multirow{2}{*}{0.275}} & BM25 & 0.0001 &  & CLIP & 0.022 \\
 & \multicolumn{1}{l|}{} & BGE-m3 & 0.131 &  & SigLip & 0.109 \\ 
\multirow{2}{*}{PDF Parser} & \multicolumn{1}{l|}{\multirow{2}{*}{0.762}} & BGE-large & 0.137 &  & Col-Paligemma & 0.140 \\
 & \multicolumn{1}{l|}{} & NV-embed-v2 & 0.117 &  & Col-Phi-3-V & 0.230 \\ \cline{1-4}
 &  &  &  &  & Col-InternVL2 & 0.581 \\ \cline{6-7} 
\end{tabular}
}
\caption{Per-page time cost of retrieval methods: The left table presents time cost (seconds) of text-based methods that rely on text extraction techniques, such as OCR models, followed by text encoders to compute page embeddings. The right table presents time cost (seconds) of multi-modal methods that encode the entire page as an image.}
\label{tab:retrieval_time}
\end{table}

\section{Additional Experiment results}
We compare our method with text-only baselines using a document parser\footnote{Adobe Extract API: \textcolor{magenta}{\href{https://developer.adobe.com/document-services/docs/overview/pdf-extract-api/}{https://developer.adobe.com/document-services/apis/pdf-extract/}}} to highlight the advantages of MLLMs in multi-modal understanding. QA results are reported for the \dataname-Bench and MMLongBench-Doc datasets, where PDF files are available.

Table \ref{tab:parser_results} presents QA results on \dataname-Bench and MMLongBench-Doc datasets. To evaluate answer quality for \dataname-Bench, where true answers are long and detailed, we introduce the Mean GPT Score (MGS), as string-matching methods often penalize variations in wording for lengthy answers. Instead, we prompt GPT-4o to compare a model’s answer with the ground truth and assign a binary score based on detail alignment.
\begin{table}[!h]
\centering
\fontsize{9}{11}\selectfont
\setlength\arrayrulewidth{0.6pt}
\resizebox{0.75\textwidth}{!}{
\begin{tabular}{lccccc}
\hline
    QA Module & Retrieval Module & Evidence & \dataname-B & MMLong \\ 
     & & & MGS & G-Acc \\
    \hline
    \multicolumn{2}{l}{\textcolor{gray}{\fontsize{7}{8}\textit{Text only QA methods}}} && \\
    Phi-3 + parser & Col-Phi-3-V & R5 & 14.1 & 29.2 \\ 
    GPT-4o + parser  & Col-Phi-3-V & R5 & 24.9 & 43.2 \\
    GPT-4o + parser & - & A & 27.6 & 42.4 \\ 
    \hline
    \multicolumn{2}{l}{\textcolor{gray}{\fontsize{7}{8}\textit{MLLM QA models}}} && \\
    PaliGemma  & Col-PaliGemma & R1 & 12.2 & 23.9  \\ 
    Phi-3-V  & Col-Phi-3-V & R1 & 24.2 & 30.7 \\ 
    \papername-InternVL2 &  Col-InternVL2 &   R5 & 25.2 & 33.2 \\ 
    GPT-4o &  Col-Phi-3-V & R5 & 47.2 & 55.1 \\ 
    GPT-4o &  - & A & 43.2 & 54.5 \\ \hline
\end{tabular}
}
\caption{parser results}
\label{tab:parser_results}
\end{table}

Our results indicate that using image evidence consistently outperforms text-only evidence. On \dataname-Bench, text-only baselines showed a significant performance drop, emphasizing the dependency of questions on both image and text. However, the MGS of text-only baselines is not zero, likely because the model leverages text from a broader context rather than relying solely on the surrounding text, enabling it to extract relevant information even in the absence of visual input.

Additionally, reducing input pages with the retrieval module improved GPT-4o’s performance with image evidence, aligning with the findings in Table \ref{tab:sp_results}. In contrast, retrieval did not enhance GPT-4o’s performance on \dataname-Bench in the text-only setting, likely because the evidence pages lacked sufficient information to fully address the questions. Including additional context in such cases might yield better results.